\documentclass{amsart}
\usepackage[foot]{amsaddr}

\usepackage[numbers]{natbib}

\usepackage[toc,page]{appendix}

\usepackage[utf8]{inputenc} 
\usepackage[T1]{fontenc}    
\usepackage{hyperref}       
\usepackage{url}            
\usepackage{booktabs}       
\usepackage{amsfonts,amsmath,amsthm}       
\usepackage{nicefrac}       
\usepackage{microtype}      
\usepackage{etex}           
\usepackage[pdftex]{graphicx}
\usepackage{enumerate}      
\usepackage{stmaryrd}
\usepackage{proof}
\usepackage{tablefootnote}

\usepackage{clipboard}

\usepackage{subfig}

\usepackage{xparse}
\usepackage{soul}

\usepackage[dvipsnames]{xcolor}
\definecolor{lightgray}{RGB}{192 192 192}
\definecolor{todocolor}{RGB}{158 90 76}
\definecolor{unused}{RGB}{32,144,140}
\definecolor{positive}{RGB}{253,231,36}
\definecolor{negative}{RGB}{68,1,84}
\definecolor{falsecolor}{RGB}{140,80,40}
\definecolor{truecolor}{RGB}{218,112,214}
\definecolor{leftcolor}{RGB}{0,171,255}
\definecolor{rightcolor}{RGB}{226,44,214}
\definecolor{bothcolor}{RGB}{241,114,36}
\usepackage{tikz-cd}
\usepackage{tikz}
\usetikzlibrary{decorations.pathmorphing,shapes}

\usepackage{float}
\usepackage{wrapfig}

\usepackage{algorithm}
\usepackage{algpseudocode}
\algblock{ParFor}{EndParFor}
\algnewcommand\algorithmicparfor{\textbf{parfor}}
\algnewcommand\algorithmicpardo{\textbf{do}}
\algnewcommand\algorithmicendparfor{\textbf{end\ parfor}}
\algrenewtext{ParFor}[1]{\algorithmicparfor\ #1\ \algorithmicpardo}
\algrenewtext{EndParFor}{\algorithmicendparfor}

\newcommand{\x}{\ensuremath{\times}}

\RenewDocumentCommand{\to}{E{^_}{{}{}}}{ \xrightarrow[\,{#2}\,]{\,{#1}\,} }

\newtheorem{proposition}{Proposition}
\newtheorem*{proposition*}{Proposition}
\newtheorem*{lemma*}{Lemma}

\newtheorem{lemma}{Lemma}

\sethlcolor{yellow}

\title[Self-SATisfied]{Self-SATisfied: An End-to-End Framework for SAT Generation and Prediction}

\author{Christopher R. Serrano}
\author{Jonathan Gallagher}
\author{Kenji Yamada}
\author{Alexei Kopylov}
\author{Michael A. Warren}
\address{All authors were at HRL Laboratories, LLC , Malibu, CA 90265
  when this work was completed.  Serrano is currently at Lockheed
Martin Corporation.}
\email{jgallagher@hrl.com, kyamada@hrl.com, mawarren@hrl.com}

\begin{document}

\maketitle

\begin{abstract}
  The boolean satisfiability (SAT) problem asks whether there exists an assignment
  of boolean values to the variables of an arbitrary boolean formula making the
  formula evaluate to True. It is well-known that all NP-problems can be coded as
  SAT problems and therefore SAT is important both practically and theoretically.
  From both of these perspectives, better understanding the patterns and structure implicit in
  SAT data is of significant value. In this paper, we describe several advances
  that we believe will help open the door to such understanding: we introduce
  hardware accelerated algorithms for fast SAT problem generation, a geometric SAT
  encoding that enables the use of transformer architectures typically applied to
  vision tasks, and a simple yet effective technique we term \emph{head slicing}
  for reducing sequence length representation inside transformer architectures.
  These advances allow us to scale our approach to SAT problems with thousands of
  variables and tens of thousands of clauses. We validate our architecture, termed
  \emph{Satisfiability Transformer (SaT)}, on the SAT prediction task with data
  from the SAT Competition (SATComp) 2022 problem sets. Prior related work either
  leveraged a pure machine learning approach, but could not handle SATComp-sized
  problems, or was hybrid in the sense of integrating a machine learning component
  in a standard SAT solving tool. Our pure machine learning approach achieves
  prediction accuracies comparable to recent work, but on problems that are an
  order of magnitude larger than previously demonstrated. A fundamental aspect of
  our work concerns the very nature of SAT data and its suitability for training
  machine learning models.
\end{abstract}

\section{Introduction}

Boolean satisfiability (SAT) is a paradigmatic example of
an NP-complete problem \cite{Cook1971TheCO} that has been widely
studied for both its theoretical importance in complexity theory and
its practical applications. From the perspective of machine learning,
SAT also presents a compelling challenge. Even in its simplest form as
a learning problem, so-called \emph{SAT prediction}, which amounts to a binary
classification task, extraordinary richness of the data presents
a number of intriguing technical obstacles to be overcome. First,
because SAT is NP-complete, generating labeled data for hard problem
instances is non-trivial. Indeed, it is well known in the complexity theory
literature that an efficient generator for guaranteed
hard problems would yield a proof that $\mathsf{P}=\mathsf{NP}$ (see e.g. \cite{article:levin-oneway,article:impagliazzo-five-worlds}).
Second, finding ways to
scale SAT prediction to large problems, of the kind typically
encountered in industry, using a pure machine learning approach
requires a careful choice of architecture.

Inspired by the pioneering work of NeuroSAT \cite{Selsam2019LearningAS},
we introduce a pure machine learning approach to both SAT problem
generation and SAT prediction. Our approach is novel in several
respects. First, our SAT data generation algorithms are designed to
run on the GPU and therefore enable end-to-end data generation and
training. This is crucial in order to train a SAT predictor at
scale. Second, by representing SAT problem data as images and adapting
the Vision Transformer (ViT) architecture of \cite{Dosovitskiy2021AnII} with a novel
sequence length reduction mechanism, we obtain an efficient architecture, called
the Satisfiability Transformer (SaT), that scales to orders of magnitude
larger problems than the prior work on pure machine learning SAT
prediction, all on a single GPU. Finally, because we are able to
tackle larger problems, we are able to validate not only against data
generated in the same manner as our training data, as was the case for
prior work on pure machine learning approaches, but also against
validation data from the industry standard (for traditional,
non-machine learning or hybrid algorithms) SATComp competition
datasets. This allows us to better understand the generalizability of
our algorithms and opens the door to further future exploration of SAT
and related problems using a pure machine learning approach.

\section{Background and Related work}\label{sec:related-work}

\subsection{Boolean Satisfiability}
In this section we define the problems SAT and UNSAT, describe why
generating instances, consisting of a problem together with its correct label, of each
might appear hard, why generating \emph{hard} instances of each is hard, and previous work that
creates datasets of SAT and UNSAT instances.

Any expression built from the constants $0,1$, the operations $\vee,\wedge,\neg$ and variables is a
\textbf{boolean formula}.
A boolean formula $\varphi(x_1,\ldots,x_n)$ is \textbf{satisfiable} in case there are choices $a_1,\ldots,a_n \in \{0,1\}$,
also known as a variable \textbf{assignment},
such that $\varphi(a_1,\ldots,a_n) = 1$.  For example $\varphi(x,y) := x \vee \neg y$ is satisfiable since $(1,1)$
is a witness in the sense that $1\vee \neg 1 = 1$.  A formula $\varphi(x_1,\ldots,x_n)$ is \textbf{unsatisfiable} in
case for every choice $a_1,\ldots,a_n$, $\varphi(a_1,\ldots,a_n)=0$.  For example $\varphi(x) := \neg(\neg x \vee \neg \neg x)$
is unsatisfiable.

Boolean formulae are often represented in a simplified form known as \textbf{conjunctive normal
form (CNF)}. CNF formulae have the form $\bigwedge_{i=1}^m (\bigvee_{j=1}^{m_i}l_{ij})$
where each so-called \textbf{literal} $l_{ij}$ is either a variable or the negation of a
variable. The disjunctions are called \textbf{clauses}
(resp. $k$-clauses when $m_i=k$). This is a $k$-\textbf{CNF} if each
$m_i=k$.
The Tseitin transformation \cite{article:tseitin-transform} associates a CNF
to any boolean formula with only polynomial overhead in a way that preserves
satisfiability.  The SAT classification task is the classification task of determining
whether an arbitrary CNF is satisfiable or unsatisfiable \footnote{In the literature the SAT and CNF-SAT tasks
are sometimes distinguished.}, and is also the task we wish to predict using machine learning.

SAT is a well known NP-complete problem \cite{Cook1971TheCO}: given a CNF $\varphi$ and an assignment $\alpha$,
there is a polynomial time algorithm that checks if $\varphi(\alpha) =
1$ and any NP problem can be transformed into SAT in polynomial time.  UNSAT is a well known co-NP-complete problem: this means that given a
CNF $\varphi$ and an assignment $\alpha$, there is a polynomial time algorithm that checks if $\alpha$ is a counterexample
to purported unsatisfiability of $\varphi$ i.e. $\varphi(\alpha)=1$, and that the complement of any NP problem
can be transformed into UNSAT in polynomial time.  It is known that $\mathsf{NP} \supseteq \mathsf{P} \subseteq \mathsf{co\mbox{-}NP}$;
however, it is not known if either of the inclusions are strict nor is it known if $\mathsf{NP}$ and $\mathsf{co\mbox{-}NP}$
are distinct, and the challenge to characterize the strictness or lack
thereof is widely regarded as one of the most important unsolved
problems in computer science and mathematics.

The formulation of NP and co-NP problems only provide algorithms for finding a variable assignment $\alpha$
in the sense that given the algorithm for checking $\varphi(\alpha)=1$, one can brute-force search all the (exponentially many) such
$\alpha$.  However, more sophisticated algorithms have been designed that, with worst-case complexity still exponential,
are often efficient, especially on real-world instances.
\textbf{Resolution} uses a proof calculus for boolean logic with one
rule: resolution.  Resolution states that from CNF clauses $l_1 \vee
\cdots \vee l_k\vee l_{k+1}$ and $u_1\vee \cdots \vee u_j\vee \neg
l_{k+1}$, the new clause $l_1\vee \ldots\vee l_k\vee u_1\vee \ldots
u_j$ can be inferred (the case $(b \vee \neg a) \wedge a$ yields the
rule modus ponens).
Resolution is \emph{complete} in that any unsatisfiable formula can be proved unsatisfiable by finding a sequence of
resolution deductions that eventually derives the empty clause.  The resolution proof search algorithm applies resolution
until resolution either derives the empty clause (in which case the formula is unsatisfiable) or any application of
the resolution rule fails to produce a new clause.  In the latter case, the formula is satisfiable, and indeed an
assignment can be read off \cite{article:dp-resolution}.

One goal in creating a dataset for use in training a machine learning
classifier is that the dataset should contain some hard problems; otherwise, the dataset may cause the
classifier to learn a quirk for classifying easy problems.  However, it is known that any algorithm which could produce always hard SAT or UNSAT problems
in polynomial time yields a \emph{one-way function}, and would serve as a separator of $\mathsf{P}$ and $\mathsf{NP}$ \cite{article:impagliazzo-five-worlds,article:levin-oneway}.
This makes it unlikely that a dataset can be produced for SAT or UNSAT that can always produce hard problems.
The various computational hardness properties have led researchers to address creating datasets for training
machine learning classifiers in the past using a variety of clever techniques, each of which attempts to
address a desired property of the resulting dataset.

\subsection{Applications of machine learning to SAT}

One of the first and most prominent applications of neural networks to
SAT prediction was NeuroSAT \cite{Selsam2019LearningAS}, which used a graph neural network (GNN)
architecture together with a CPU-bound data generation method that
leverages an external SAT solver. Unlike NeuroSAT, we use a transformer
architecture and a novel data generation method that is not tied to an
external solver. Earlier work \cite{Devlin08satisfiabilityas} similarly formulates SAT as a
machine learning classification problem, though unlike our methodology it requires
hand-crafted features extracted by the SATzilla \cite{10.1007/978-3-540-74970-7_50}
algorithm portfolio for SAT solving. Our approach, which learns features and
heuristics directly from the data, automates the crafting of problem features
and heuristics removing the need for experts in the field to invent such
heuristics by hand. Other work such as NeuroCore
\cite{10.1007/978-3-030-24258-9_24} and NeuroComb
\cite{Wang2021NeuroCombIS} present hybrid approaches combining neural
networks with traditional SAT solving algorithms. While these
approaches are of interest, we are here concerned with the pure
machine learning approach and the SAT prediction (not SAT solving) problem.

Recent related work is that of Polu and
Sutskever \cite{Polu2020GenerativeLM} in which a transformer language
model \cite{Radford2019LanguageMA} is applied to the task of automated theorem
proving in the Metamath \cite{metamath} formal environment. However, this work
is focused on tactic-level, as opposed to low-level (our case),
theorem proving and therefore does not apply to the kinds of
constraint satisfaction tasks we address.
Recent work similarly applies transformer architectures, in
combination with a GNN embedding, to SAT
prediction \cite{Shi2022SATformerTF}. However, the approach in \emph{ibid.}
is limited to problems with a maximum of 60 variables, whereas we demonstrate
our approach on problems with 1,500 variables. Further, we validate our
performance against a subset of SATComp problems, as opposed to validating
against the same generator that generated our training set (as in \emph{ibid.}),
and achieve comparable accuracy scores (60.38\% on SATComp problems up to 1500
variables (ours) vs. 61\% on problems up to 60 variables (theirs)).

\subsection{Transformer architectures for input sequence length reduction}\label{sec:related-work-model-architecture}
A transformer block typically maps a sequence of $N$ input tokens to $N$ output
tokens. The self-attention portion of the transformer block performs pairwise
comparisons between the tokens of the input sequence, requiring $O(N^2)$ memory
and time complexity. This computational bottleneck has spawned research into
techniques that reduce the length of the output sequence, either by modifying
the self-attention algorithm directly \cite{Wu2021CentroidTL,
Dai2020FunnelTransformerFO}, via a learned heuristic for down-sampling output
tokens \cite{Huang2022PyramidBERTRC, Kim2020LengthAdaptiveTT}, or via
down-sampling policies learned through evolutionary
algorithms \cite{Kim2020LengthAdaptiveTT} or reinforcement
learning \cite{Ye2021TRBERTDT}. Convolutional neural network (CNN) based vision
architectures have long made use of pooling to down-sample feature
maps \cite{Gholamalinezhad2020PoolingMI}, and similar approaches have been
applied to Vision Transformer architectures to reduce patch
size \cite{Heo2021RethinkingSD, Pietruszka2020SparsifyingTM}, including to
reduce the sequence length \cite{Pan2021ScalableVT,Wang2021PyramidVT}. These
approaches, analogous to pooling in CNNs, rely on the notion of locality
inherent in image processing to down-sample by combining local features. Our work
makes no assumptions of locality and does not rely on notions of spatial
relationships in the input sequence, making it more general.

\section{Methods}

\subsection{Variable space encoding}\label{section:sat-as-vision}

We represent boolean formula in a geometric encoding we term the \emph{variable
space encoding} that allows us to relate boolean formulae to a computer vision
problem and leverage recent developments in Vision Transformer architectures.
We canonically represent any CNF $\varphi$ in $v$ variables
and $m$ clauses as an $m\x v$ matrix, $\underline{\varphi}$, over $\{-1,0,1\}$.
$\underline{\varphi}_{i,j}$ is $0$ if $x_j$ does not appear in the $i^\text{th}$
clause, is $1$ if it appears positively and is $-1$ if it appears negatively.

Similar encodings of CNF have appeared frequently in the literature. If one
turns all the occurrences of $-1$ to $1$, then $\underline{\varphi}$ is the
adjacency matrix for the bipartite literal incidence graph used in
\cite{DBLP:journals/corr/abs-2302-00272,conference:satgraf}. It has also been directly studied as the labeled adjacency matrix for
the bipartite factor graph of the CNF \cite{conference:universalsatgraph} \footnote{A CNF
  presentation of a formula $\varphi$ may be regarded as a completely
  multiplicatively factored presentation of a polynomial over the two element
ring.}. The novelty in our approach to this encoding is to regard it
geometrically, by associating an image to such a matrix where each of $-1,0,1$
is mapped to a distinct color, turning each cell in the $m\x v$ into a pixel.
We can thus bring state-of-the-art Transformer-based Computer
Vision models \cite{Dosovitskiy2021AnII} to the task of predicting SAT/UNSAT by
treating each row or column as a \emph{patch} of the ``image''. It also gives rise
to a visual presentation of the meaning of satisfiability.

\begin{figure}
  \centering
  \includegraphics[width=0.45\textwidth]{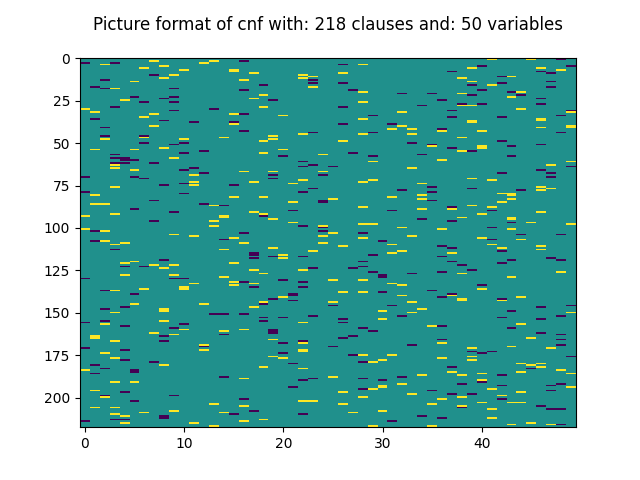}
  \caption{Visualization of variable space encoding.}
\end{figure}

\subsection{Training Data: Labeled CNF generation}\label{sec:data-generation}

Creating boolean formulae is easy; however, as explained earlier, creating boolean formulae with
labels indicating whether they are satisfiable or unsatisfiable, comes with fundamental challenges.  One cannot rely
on generating formulae and then testing their satisfiability \emph{ex post facto}.  We require generating on the
order of billions of labeled pairs of CNF, that have  broad variability, and with variables and clauses scaling to the 1k and 10k
levels respectively.  In this section, we describe our algorithms
for generating data as pairs of a CNF $\varphi$ and a label $\mathsf{SAT}$ or $\mathsf{UNSAT}$.  Our algorithms, called the \emph{SATisfactory},
run on a single GPU and are highly parallel, enabling the required scaling.

\begin{figure}
  \begin{tikzpicture}[scale=0.2]
    \begin{scope}[local bounding box=L]
        \fill[positive] (0,1) rectangle ++ (1,1); 
        \fill[negative] (1,1) rectangle ++ (1,1); 
        \fill[negative] (2,1) rectangle ++ (1,1); 
        \node at (4.25,1.5) {$\cdots$}; 
        \fill[positive] (5.25,1) rectangle ++ (1,1); 
        \draw (0,1) grid (3,2); 
        \draw[black] (5.25,1) rectangle ++ (1,1); 
    \end{scope}

\begin{scope}[yshift=-5.0cm,local bounding box=LL]
    \draw[truecolor,line width=0.05cm] (0,0) rectangle ++ (1,1); 
    \draw[falsecolor,line width=0.05cm] (1,0) rectangle ++ (1,1); 
    \draw[black] (2,0) rectangle ++ (1,1); 
    \draw[falsecolor,line width=0.05cm] (5.25,0) rectangle ++ (1,1) ; 
    \node[align=center] at (1.5,1.5) {$\cdots$} ; 
    \draw[falsecolor,line width=0.05cm] (0,2) rectangle ++ (1,1) ; 
    \draw[truecolor, line width=0.05cm] (1,2) rectangle ++ (1,1) ; 
    \draw[truecolor, line width=0.05cm] (2,2) rectangle ++ (1,1) ; 
    \draw[black] (5.25,2) rectangle ++ (1,1) ; 
    \node[align=center] at (4.25,2.5) {$\cdots$}; 
    \draw[black] (0,3) rectangle ++ (1,1) ; 
    \draw[truecolor, line width=0.05cm] (1,3) rectangle ++ (1,1) ; 
    \draw[black] (2,3) rectangle ++ (1,1); 
    \draw[truecolor, line width=0.05cm] (5.25,3) rectangle ++ (1,1); 
    \draw[gray] (-0.25,-0.25) rectangle ++ (6.75,4.5);
\end{scope} 

\begin{scope}[yshift=-11.0cm,local bounding box=LLL]
    \fill[positive] (0,0) rectangle ++ (1,1); 
    \draw[truecolor,line width=0.05cm] (0,0) rectangle ++ (1,1); 
    \draw[falsecolor,line width=0.05cm] (1,0) rectangle ++ (1,1); 
    \draw[black] (2,0) rectangle ++ (1,1); 
    \draw[falsecolor,line width=0.05cm] (5.25,0) rectangle ++ (1,1) ; 
    \node[align=center] at (1.5,1.5) {$\cdots$} ; 
    \draw[falsecolor,line width=0.05cm] (0,2) rectangle ++ (1,1) ; 
    \fill[negative] (1,2) rectangle ++ (1,1) ;
    \fill[negative] (2,2) rectangle ++ (1,1) ;
    \draw[truecolor,line width=0.05cm] (1,2) rectangle ++ (1,1);
    \draw[truecolor,line width=0.05cm] (2,2) rectangle ++ (1,1);
    \draw[black] (5.25,2) rectangle ++ (1,1) ; 
    \node[align=center] at (4.25,2.5) {$\cdots$}; 
    \draw[black] (0,3) rectangle ++ (1,1) ; 
    \fill[negative] (1,3) rectangle ++ (1,1) ; 
    \draw[truecolor, line width=0.05cm] (1,3) rectangle ++ (1,1) ; 
    \draw[black] (2,3) rectangle ++ (1,1); 
    \fill[positive] (5.25,3) rectangle ++ (1,1); 
    \draw[truecolor, line width=0.05cm] (5.25,3) rectangle ++ (1,1); 
    \draw[gray] (-0.25,-0.25) rectangle ++ (6.75,4.5);
\end{scope} 

\begin{scope}[yshift=-17.0cm,local bounding box=LLLL]
    \fill[positive] (0,0) rectangle ++ (1,1); 
    \fill[positive] (1,0) rectangle ++ (1,1); 
    \draw[black] (1,0) rectangle ++ (1,1);
    \draw[truecolor,line width=0.05cm] (0,0) rectangle ++ (1,1); 
    \fill[unused] (2,0) rectangle ++ (1,1); 
    \draw[black] (2,0) rectangle ++ (1,1);
    \fill[negative] (5.25,0) rectangle ++ (1,1) ; 
    \draw[black] (5.25,0) rectangle ++ (1,1);
    \node[align=center] at (1.5,1.5) {$\cdots$} ; 
    \fill[negative] (0,2) rectangle ++ (1,1) ; 
    \draw[black] (0,2) rectangle ++ (1,1);
    \fill[negative] (1,2) rectangle ++ (1,1) ;
    \fill[negative] (2,2) rectangle ++ (1,1) ;
    \draw[truecolor,line width=0.05cm] (1,2) rectangle ++ (1,1);
    \draw[truecolor,line width=0.05cm] (2,2) rectangle ++ (1,1);
    \fill[unused] (5.25,2) rectangle ++ (1,1) ; 
    \draw[black] (5.25,2) rectangle ++ (1,1);
    \node[align=center] at (4.25,2.5) {$\cdots$}; 
    \fill[unused] (0,3) rectangle ++ (1,1) ; 
    \draw[black] (0,3) rectangle ++ (1,1);
    \draw[black] (2,3) rectangle ++ (1,1);ÍÍ
    \fill[negative] (1,3) rectangle ++ (1,1) ; 
    \draw[truecolor, line width=0.05cm] (1,3) rectangle ++ (1,1) ; 
    \fill[unused] (2,3) rectangle ++ (1,1); 
    \fill[positive] (5.25,3) rectangle ++ (1,1); 
    \draw[truecolor, line width=0.05cm] (5.25,3) rectangle ++ (1,1); 
    \draw[gray] (-0.25,-0.25) rectangle ++ (6.75,4.5);
\end{scope} 


\begin{scope}[xshift=14cm,local bounding box=R]
\node[align=left] at (0,1.5) {\tiny{Select assignment}} ;
\end{scope}

\begin{scope}[xshift=14cm,yshift=-2.5cm,local bounding box=RR]
    \node[align=left] at (0,2.0) {\tiny{Select variables}} ;
    \node[align=left] at (0,.5) {\tiny{in each row that}} ;
    \node[align=left] at (0,-1) {\tiny{will \fcolorbox{white}{truecolor}{\ } and will not \fcolorbox{white}{falsecolor}{\ }}} ;
    \node[align=left] at (0,-2.5) {\tiny{be satisfied}} ;
\end{scope} 

\begin{scope}[xshift=14cm,yshift=-8.5cm,local bounding box=RRR]
    \node[align=left] at (0,1) {\tiny{Match satisfying variables }} ;
    \node[align=left] at (0,-.5) {\tiny{variables to}} ;
    \node[align=left] at (0,-2) {\tiny{assignment}} ;
\end{scope} 

\begin{scope}[xshift=14cm,yshift=-14.5cm,local bounding box=RRRR]
    \node[align=left] at (0,0) {\tiny{Fill in everything}} ;
    \node[align=left] at (0,-1.5) {\tiny{else appropriately}} ;
\end{scope} 

\end{tikzpicture}

  ~

  \input{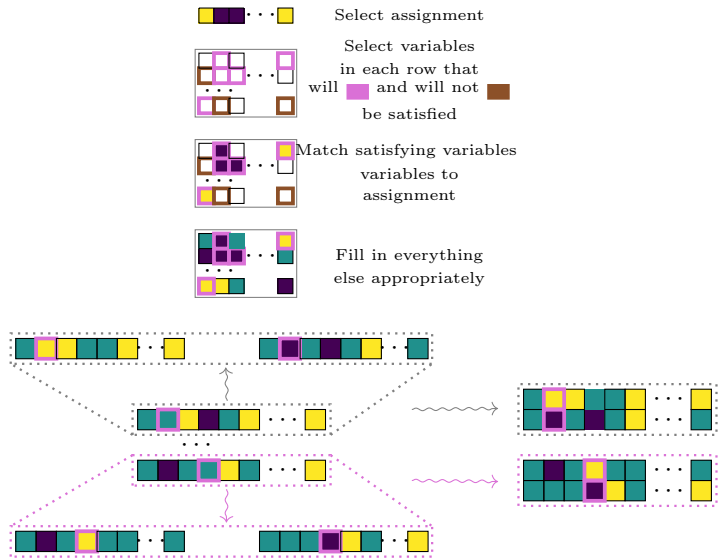}

  \caption{Depictions of (top) the SAT instance generator, where the steps of the algorithm
    run from top to bottom, and (bottom) the resolution blooming process.
  \label{fig:sat-gen-vision}}
\end{figure}

The random process by which we generate satisfiable problems follows from a tautological observation:
a satisfiable formula must have a satisfying assignment.  The intuitive view on turning this observation
into an algorithm for generating satisfiable formulae, represented in the variable
space encoding of Section \ref{section:sat-as-vision}, is shown in
Figure \ref{fig:sat-gen-vision}
and described in Algorithm \ref{algorithm:satgen}. Here
$\mathsf{vars}$ denotes a family of probability distributions on
finite ordinals and we write $\mathsf{var}(n)$
for the distribution on $\{1,\ldots,n\}$.  For example, the family of uniform distributions on $\{1,\ldots,n\}$ is such
a family.  The variable $\mathsf{lits\mbox{-}clause}$ is a distribution on $\{1,\ldots,n\}$
and is used to determine the number of literals in each clause.  The variable $\mathsf{polarities}$  is a
distribution on $\{-1,1\}$ and is used to determine whether a variable in a satisfying assignment is set to $\mathsf{false},\mathsf{true}$.
Finally, the variable $\mathsf{polarity\mbox{-}bias}(l)$ is a distribution family over binary sequences of
length $l$ that contain at least one $1$.  The distribution family
which, for any $l$, is the uniform distribution uniform on $\{1,\ldots,2^l-1\}$, as a binary string of length $l$, is an example.
\begin{algorithm}
  \caption{SATisfactory SAT Instance Generator}
  \label{algorithm:satgen}
  \begin{algorithmic}[1]
    \Function{GenerateSAT}{$n,m,\mathsf{vars},\mathsf{lits\mbox{-}clause},\mathsf{polarities},\mathsf{polarity\mbox{-}bias}$}
    \State $\mathsf{asst} \gets$ sample $n$ times independently from $\mathsf{polarities}$
    \State Initialize $\mathsf{prob}$ as a CNF in $n$ variables with $m$ clauses
    \For{each of the $m$ clauses $c_i$}
    \State $\mathsf{num\mbox{-}lits} \gets $ sample the number of literals in $c_i$ from $\mathsf{lits\mbox{-}clause}$
    \State $\mathsf{var\mbox{-}indices} \gets $ sample $\mathsf{num\mbox{-}lits}$ indices independently, uniquely  from $\mathsf{vars}(n)$.
    \For{$j \in (1,\mathsf{num\mbox{-}lits})$ and $v_j \gets \mathsf{var\mbox{-}indices}[j]$}
    \State $\mathsf{seq} \gets$ sample one sequence from $\mathsf{polarity\mbox{-}bias}(\mathsf{num\mbox{-}lits})$.
    \State $c_i[v_j] \gets \mathsf{asst}[v_j]$ in case $\mathsf{seq}[j] == 1$ otherwise $c_i[v_j] \gets -\mathsf{asst}[v_j]$
    \EndFor
    \EndFor
    \State \Return $\mathsf{prob}$
    \EndFunction
  \end{algorithmic}
\end{algorithm}

We now state a theoretical guarantee on this algorithm, the
proof of which can be found in Appendix \ref{appendix:proof:sat-soundness}.
\begin{proposition}[Soundness]\label{proposition:sat-soundness}
  \Copy{proposition:sat-soundness}{The generator from Algorithm \ref{algorithm:satgen} produces only
  satisfiable formulae.}
\end{proposition}
With the above note that we can eventually generate any $k$-clause that satisfies an assignment, and also that
we will eventually generate all assignments.  We thus have a weak completeness result in that sense.
Depending on the distribution $\mathsf{polarity\mbox{-}bias}$, a more nuanced version of this completeness result may be
developed, but is outside of the scope of this paper.

The random process by which we generate unsatisfiable problems
follows from the observation that a formula $\varphi$ is
unsatisfiable iff one can form a chain of implications
$\varphi_{1} \Rightarrow \cdots \Rightarrow \varphi_{n} \Rightarrow \bot$. We then start with
$\bot$ and randomly branch backwards using resolution to create, e.g.,
\[
  \bot
  \Leftarrow  x_{i_1} \wedge \neg x_{i_1}
  \Leftarrow (x_{i_1} \vee x_{i_2}) \wedge (x_{i_1}\vee\neg x_{i_2}) \wedge(\neg x_{i_1} \vee x_{i_3}) \wedge (\neg x_{i_1} \vee \neg x_{i_3})
  \Leftarrow \cdots
\]
We call the process of randomly constructing resolution proof trees \emph{blooming} due to a strong connection to stochastic L-systems (see \cite{LINDENMAYER1968300,413300}).
Where in stochastic L-systems trees are created by randomly applying stochastic grammar production rules in parallel,
we create trees by randomly applying the resolution rule, in the proof search direction, in parallel.  The blooming process is simple to describe directly,
is visualized in
Figure \ref{fig:sat-gen-vision}
and is formalized in Algorithm \ref{algorithm:unsatgen}.
\begin{algorithm}
  \caption{SATisfactory UNSAT Instance Generator}
  \label{algorithm:unsatgen}
  \begin{algorithmic}[1]
    \Function{ResSearch}{$\mathsf{prob},\mathsf{bloom}$} \Comment $\mathsf{prob}$ has $m$ clauses and $n$ variables
    \State Initialize a CNF $\mathsf{new\mbox{-}prob}$ with $2m$ clauses and $n$ variables
    \For{each of the $m$ clauses of $\mathsf{prob}$, $c_i$}
    \State $\mathsf{available}_i \gets$ unused variables in $c_i$
    \State $\mathsf{unavailable}_i \gets $ used variables in $c_i$
    \State $\mathsf{cut\mbox{-}var} \gets$ sample $1$ variable index from $1,\mathsf{len}(\mathsf{available}_i)$ from the distribution $\mathsf{var}$
    \State $\mathsf{new\mbox{-}prob}[2i,\mathsf{cut\mbox{-}var}] \gets 1$
    \State $\mathsf{new\mbox{-}prob}[2i-1,\mathsf{cut\mbox{-}var}] \gets -1$
    \State $\mathsf{bloom\mbox{-}choices} \gets$ sample $\mathsf{len}(\mathsf{unavailable_i})$ selections from $\{0,1,2\}$ from $\mathsf{bloom}$
    \For{each of the $\mathsf{bloom\mbox{-}choices}$, $\mathsf{ch}_k$}
    \State $\mathsf{new\mbox{-}prob}[2i,\mathsf{unavailable}_i[k]]
    \gets {
      \begin{cases}
        \mathsf{prob}[i,\mathsf{unavailable}_i[k]] & \mathsf{ch}_k==0,2\\
        0 & \text{otherwise}
    \end{cases}}$
    \State $\mathsf{new\mbox{-}prob}[2i\mbox{-}1,\mathsf{unavailable}_i[k]]
    \gets {
      \begin{cases}
        \mathsf{prob}[i-1,\mathsf{unavailable}_i[k]] & \mathsf{ch}_k==1,2\\
        0 & \text{otherwise}
    \end{cases}}$
    \EndFor
    \EndFor
    \State \Return $\mathsf{new\mbox{-}prob}$
    \EndFunction
    ~\\
    \Function{GenerateUNSAT}{$n,m,\mathsf{init\mbox{-}size},\mathsf{depth},\mathsf{down\mbox{-}clause},\mathsf{vars},\mathsf{lits\mbox{-}clause},\mathsf{polarities},\mathsf{bloom}$}
    \State Initialize a CNF, $\mathsf{prob}$, with $2\cdot\mathsf{init\mbox{-}size}$ clauses and $n$ variables
    \For{$i \in (1,\mathsf{init\mbox{-}size})$}
    \State $j \gets $ sample $1$ index from $\mathsf{var}(1...n)$
    \State $\mathsf{prob}[2i,j] \gets 1$
    \State $\mathsf{prob}[2i-1,j] \gets -1$
    \EndFor
    \For{$\mathsf{depth}$ iterations}
    \If{$2\cdot \mathsf{prob}.\mathsf{shape}[0] > m$ or some clause has used all $n$ variables}
    \State $\mathsf{break}$
    \EndIf
    \State $\mathsf{prob\mbox{-}r\mbox{-}idcs} \gets \mathsf{down\mbox{-}clause}$
    \State $\mathsf{prob}_{rs} \gets \mathsf{prob}[\mathsf{prob\mbox{-}rs\mbox{-}idcs}]$
    \State $\mathsf{prob}_{lo} \gets \mathsf{prob}[\mathsf{prob\mbox{-}rs\mbox{-}idcs}^c]$
    \State $\mathsf{prob}_{rs} \gets \mathsf{ResSearch}(\mathsf{prob}_{rs},\mathsf{bloom})$
    \State $\mathsf{prob} \gets \mathsf{prob}_{rs} \wedge \mathsf{prob}_{lo}$
    \EndFor
    \State $\mathsf{clauses\mbox{-}needed} \gets \mathsf{max}(0,m-\mathsf{prob}.\mathsf{shape}[0])$
    \State Generate a CNF $\mathsf{r\mbox{-}prob}$ by sampling $\mathsf{clauses\mbox{-}needed}$ clauses from the distribution of clauses with variable distribution $\mathsf{vars}$, literals-per-clause distribution $\mathsf{lits\mbox{-}per\mbox{-}clause}$ and polarity distribution $\mathsf{polarities}$.
    \State $\mathsf{prob} \gets \mathsf{prob} \wedge \mathsf{r\mbox{-}prob}$.
    \State \Return $\mathsf{prob}$
    \EndFunction
  \end{algorithmic}
\end{algorithm}

The variable $\mathsf{vars}$ is a distribution family and $\mathsf{lit\mbox{-}clause},\mathsf{polarities}$
are distributions just as in the description of Algorithm \ref{algorithm:satgen}.  $\mathsf{bloom}$ is a distribution over
$\{0,1,2\}$ and is used to determine, in the $\mathsf{\sc ResSearch}$ function, whether an already occurring variable in a clause
is carried to the left subtree, the right subtree, or both.  Finally, $\mathsf{down\mbox{-}clause}$ is a distribution
over subformulae selections, so that we do not have to bloom on every single clause at every step.  We denote by $\mathsf{prob}[\mathsf{prob\mbox{-}rs\mbox{-}idcs}^c]$
those clauses not selected by $\mathsf{prob\mbox{-}rs\mbox{-}idcs}$.

In real-world problems, often, there is a small
unsatisfiable \emph{core} problem,
which may be hidden among the clauses.
We thus do not require blooming completely until the desired size, but instead allow blooming to some
depth to obtain an unsatisfiable $\varphi$ and then we generate a $\psi$ randomly, and return
$\varphi \wedge \psi$ since if $\varphi$ is unsatisfiable then so is $\varphi \wedge \gamma$ for any $\gamma$.

The following proposition, the proof of which can be found in
Appendix \ref{appendix:proof:unsat-soundness},  establishes the correctness of this generator.
\begin{proposition}[Soundness]\label{proposition:unsat-soundness}
  \Copy{proposition:unsat-soundness}{The UNSAT generator described in Algorithm \ref{algorithm:unsatgen} yields only unsatisfiable formulae.}
\end{proposition}

Like for the satisfiable instance generator, the unsatisfiable instance generator has a completeness property
with respect to finite depth resolution trees; however, the development of this property is outside of the scope of this write-up.

We also introduce some slight variations of the above algorithms where Algorithm \ref{algorithm:satgen}
can be used to transform a formula into a satisfiable one and Algorithm \ref{algorithm:unsatgen}
can be used to transform a formula into an unsatisfiable one.  The use of this technique is detailed
in the Supplemental Material.

\subsection{Satisfiability Transformer (SaT)}\label{section:SaT}

\begin{figure}
  \centering
  \includegraphics[width=.88\linewidth]{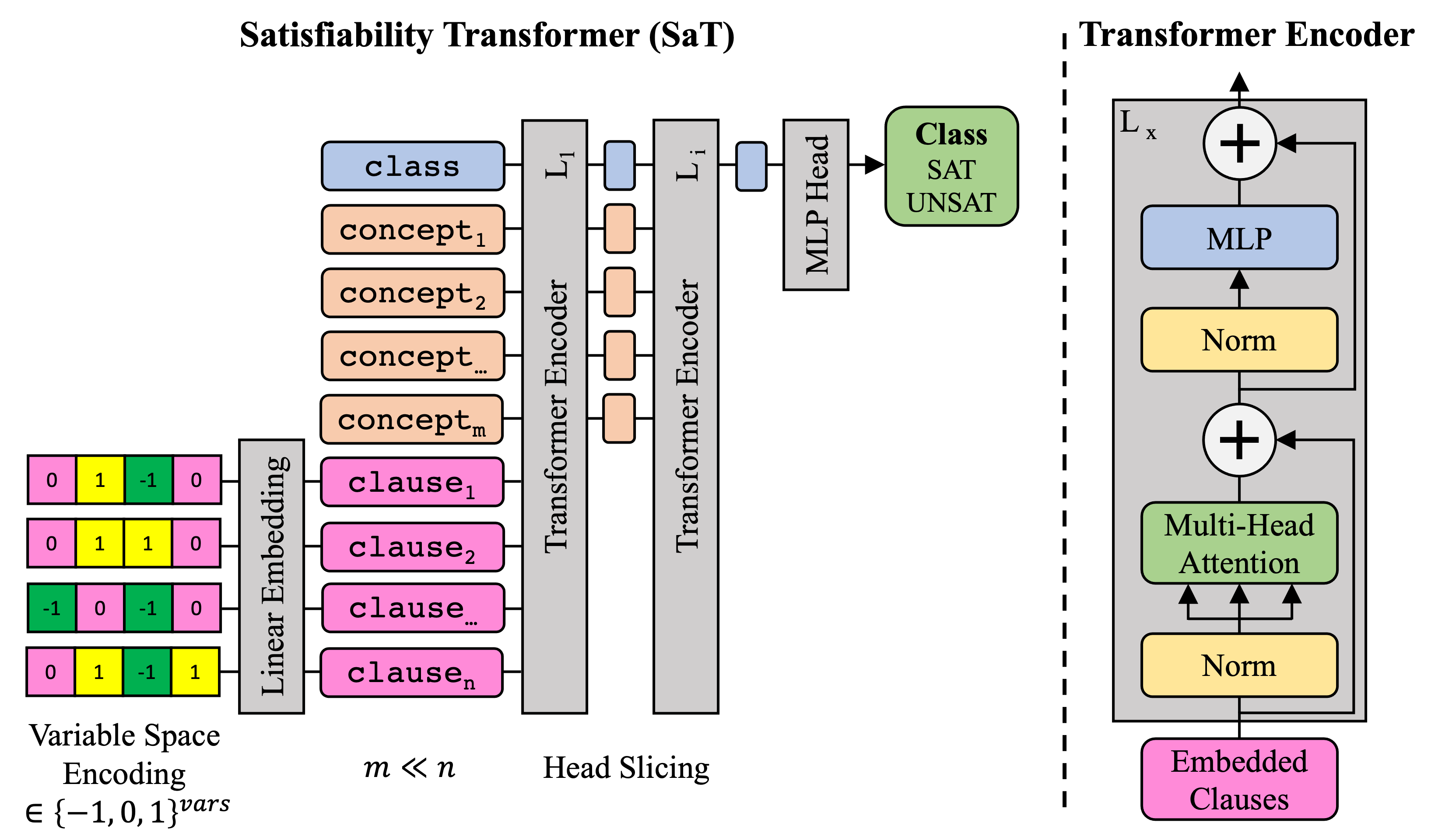}
  \caption{Model overview. We input a CNF formula in the variable space encoding
    (cf. Section \ref{section:sat-as-vision}) row-wise or column-wise. After
    embedding we prepend a \texttt{[class]} token and $m$-many learnable
    \texttt{[concept]} tokens, which we refer to as the \emph{head} of the
    sequence. After the first layer we perform head slicing (cf. Section
    \ref{section:head-slicing}) to reduce the sequence to a length of $m$+1. As in
  ViT, the \texttt{[class]} token is utilized for classification.}
  \label{fig:model-overview}
\end{figure}

We develop the Satisfiability Transformer (SaT) with inspiration from the Vision
Transformer (ViT) \cite{Dosovitskiy2021AnII}, and adopt a similar notion of
interpreting a \emph{patch} of an image as a token. However, rather than
embedding a $16\times16$ RGB patch as a token, we use a row of our variable
space encoding (cf. Section \ref{section:sat-as-vision}), which corresponds to
either a clause of the original CNF format, or the literals of a variable and
their clause membership when transposed. We leverage the natural equivariance of
self-attention by omitting the addition of any positional information to the
input. Critically, we wish to tackle SAT problems in thousands of variables,
which naturally yield tens of thousands of clauses.  Given that in our input
representation a problem of $m$ clauses and $v$ variables is represented by an
$m\times v$ image, and we take each row of the image as an input patch, the
resulting sequence length $l$ will be equal either to the number of clauses $m$
or the number of variables $v$ when transposed. We require that our model run on
a single workstation GPU, and in this work we are limited to an NVIDIA Tesla
V100S with 32GB of memory. Thus, na{\"i}ve applications of Transformer
architectures to input sequences with lengths in the thousands, given the
$O(N^2)$ complexity of self-attention, will quickly run out of memory. This
memory constraint motivates the development of a modification to the standard
transformer architecture to reduce the sequence length of the problem
representation within the model that we term \emph{head slicing}. This technique
is discussed in detail below. An overview of the Satisfiability Transformer
architecture is depicted in Figure \ref{fig:model-overview}.

\subsection{Head slicing}\label{section:head-slicing}

Inspired by the \texttt{[class]} token of BERT \cite{Devlin2019BERTPO} and
ViT \cite{Dosovitskiy2021AnII}, we additionally prepend a set of learnable
\texttt{[concept]} tokens to each input sequence $\mathbf{x}$ that we utilize for
down-sampling in subsequent layers. Formally, for an input sequence $\mathbf{x} =
(x_1, x_2, ..., x_n)$ where each $x_i$ is a vector embedding of dimension $d$
such that $x_i\in\mathbb{R}^d$, we prepend a set $head$, containing one \texttt{[class]} token
and $m$-many learnable \texttt{[concept]} tokens $c$, $head=(\texttt{[class]}, c_1, c_2,
..., c_m) \in\mathbb{R}^d$, to produce $head \circ \mathbf{x}$.  We represent
the concatenation of two finite sequences, $a$ and $b$, using the symbol
``$\circ$''.  Specifically, we denote the concatenation of sequences $a$ and $b$
as $a \circ b$. The input to the first transformer block of the encoder is thus
the concatenated sequence $head \circ \mathbf{x}$.  After $l$-many transformer
blocks the output of the $l^{th}$ block is \emph{head sliced}, such that only
the $head$ tokens are passed as input to the $l^{th}+1$ layer. Finally, after
all of the subsequent transformer blocks in the model, the \texttt{[class]}
token is passed to an MLP head to produce the binary classification. Head
slicing can be performed multiple times to produce a pyramid-like
architecture common in the computer vision literature \cite{Zhao2016PyramidSP,
Lin2016FeaturePN}. Unlike earlier work on transformer specific sequence
down-sampling methods (cf. Section \ref{sec:related-work-model-architecture}), we make
no effort to design or learn a heuristic or policy by which we select which
tokens to keep or make assumptions of locality as is common in feature map pooling.

\section{Results}\label{section:results}

It is common in the literature on machine learning for SAT prediction
to both train and validate on data produced by the same
generator, since prior work targets around 60 variables maximum and
there does not exist a representative corpus of
small, labeled CNF with so few variables (see
e.g. \cite{Selsam2019LearningAS,Shi2022SATformerTF}). Because we are
capable of training on much larger problems, we are able to
validate against problems from the SATComp 2022 dataset, thereby
obtaining a better assessment of how well our model generalizes and
also ensuring it has not learned to merely exploit artificial problem
features introduced by the data generator.

SATComp \cite{satcompetition2022} is a community-based benchmark event, held
annually since 2002. The problem sets can be either purely mathematical or based
on industrial applications and are utilized for the benchmarking of open-source
and commercial SAT solvers. Approaches to neural proving, such as
NeuroCore \cite{Selsam2019NeuroCoreGH,Wang2021NeuroCombIS}
used SATComp 2017, 2018, and 2021 for both training and testing.  We
emphasize that predicting SAT and improving a SAT solver with a neural
network, as done by NeuroCore, are qualitatively different tasks, so
that using SATComp for training and validating NeuroCore was
possible. When used for validation, SATComp problems are sampled uniformly with replacement
from the pool of problems at the appropriate number of variables and clauses.

It training of the Satisfiability Transformer using a NeuroSAT-like data generator, it became
clear around the 150 variable case, data generation was taking significantly longer than training.
To address this we designed our SAT
and UNSAT problem generators (cf. Section \ref{sec:data-generation}) to
generate problems in parallel \emph{on the fly}. The impact on generation time is
presented in Table \ref{table:timing-comparison}.  The impact on disk space is also profound;
our approach requires 0 disk space contrasted with the space required to store CNF in DIMACS
format such as for using a NeuroSAT-like generator; for example, storing 380k problems of 40--41 variables
each requires 1.5Gb, and we require on the scale of billions of
problems for full training. The generator provides many
probabilistically controlled variables and these are hyperparameters
of training.  Additional details on these hyperparameters are captured
in the Supplemental Material.
\begin{table}
  \begin{center}
    \caption{Timing results comparison}
    \label{table:timing-comparison}
    \smallskip
    \begin{tabular}{llll}
      \toprule
      \cmidrule(r){1-4}
      Generator   & Variables         & Max clauses & Problems per
      second
      \tablefootnote{Averaged
        over 5k each
      of SAT and UNSAT problems.} \\
      \midrule
      NeuroSAT       & 15               & $\sim 200$          & < 0.7      \\
      \textbf{SATisfactory}  & 15               & 440                 & 530      \\
      \textbf{SATisfactory}  & 1500             & 116500              & 128        \\
      \bottomrule
    \end{tabular}
  \end{center}
  Our SATisfactory generator scales well; when generating problems at two orders of magnitude larger,
  we experience only a 4.14 times slowdown, and remain orders of magnitude faster than the
  NeuroSAT generator.  These tests were all performed on the same machine, but due to having
  to run a traditional theorem prover, the NeuroSAT generator is CPU bound and hence does not
  take advantage of parallel generation of clauses on the GPU.
\end{table}

A direct comparison to the results of related work in the literature is
difficult, as there are relatively few end-to-end machine learning SAT/UNSAT
classification results. Closely related work, like NeuroSAT \cite{Selsam2019LearningAS} and SATformer
demonstrate SAT/UNSAT classification on problems up to 40 and 60 variables respectively.
They both  validate performance against problems generated by the same generation
scheme that produced their training set. Instead, we validate SaT against SATComp to
assess our generalizability and because comparison against a NeuroSAT-like generation scheme
is not computationally possible at the scale of larger problems.  However, this
adds the complication of not having a direct comparison since the accuracy is on
different datasets that do not have a known relationship in terms of hardness or learnability.
To address this we also perform validation of SaT on smaller problems, at the scale of NeuroSAT and SATformer,
using a variety of validation techniques including ``self-validation'', using the NeuroSAT data generation scheme, and the
$\mathsf{Comb}_\gamma$ scheme described above.

We can see in Table
\ref{table:results-comparison} that against SATComp 2022 on problems up to 1000
variables we achieve 61.26\% accuracy whereas SATformer achieves 61\% accuracy on
only 60 variables, problems an order of magnitude smaller.  On smaller problems
we significantly outperform SATformer's 61\% accuracy on 60 variables, and
NeuroSAT's 50\% accuracy on 60 variables \cite{Shi2022SATformerTF}, as we
achieve 75\% accuracy on problems up to 150 variables. Validation
against problems up to 40 or 60 variables is not meaningful, as SATComp 2022
only contains 10 problems this small, though we demonstrate an accuracy of
99\% against 41 variables on validation data generated by the NeuroSAT
data generator.

\begin{table}
  \begin{center}
    \caption{Validation results and comparison. Unlike SATformer and NeuroSAT, we validate against a subset of SATComp 2022
      problems, a benchmark typically used to evaluate the performance of
      state-of-the-art SAT solvers. For versions of SaT that are trained on problems
      of less than 150 variables we do not perform head slicing as the memory
    savings are minimal.}
    \label{table:results-comparison}
    \smallskip
    \begin{tabular}{llllll}
      \toprule
      \cmidrule(r){1-5}
      Model & Max & Max & Literals / & Validation
      &
      Accuracy \\
      & Variables & Clauses & Clause & Source & \\
      \midrule
      NeuroSAT     & 40               & $\sim 200$          & 3
      & NeuroSAT
      \cite{Selsam2019LearningAS}
      &
      84.5\%
      \cite{Selsam2019LearningAS}
      \\
      SaT (ours)   & 41               & 451                 & 3
      & NeuroSAT
      \cite{Selsam2019LearningAS}
      &
      99\%,75\%
      \tablefootnote{We
        observe
        very
      different validation accuracies depending on the bias supplied by $\mathsf{polarity\mbox{-}bias}$ in Algorithm \ref{algorithm:satgen}.  This is surprising, and a fuller understanding of this phenomena is left for future work.}  \\
      NeuroSAT     & 60               & $\sim$300           & 3               & NeuroSAT \cite{Selsam2019LearningAS}                               & 50\% \cite{Shi2022SATformerTF} \\
      SATformer    & 60               & $\sim$300           & 3               & NeuroSAT \cite{Selsam2019LearningAS}                               & 61\% \cite{Shi2022SATformerTF} \\
      SaT (ours)   & 150              & 1650                & 3               & \textbf{SATComp 2022}     & 75.00\% \\
      SaT (ours)   & 300              & 3300                & mixed           & \textbf{SATComp 2022}     & 63.75\% \\
      SaT (ours)   & 1000             & 11000               & mixed           & \textbf{SATComp 2022}     & 61.26\% \\
      SaT (ours)   & \textbf{1500}    & \textbf{16500}      & mixed           & \textbf{SATComp 2022}     & 60.38\% \\
      \bottomrule
    \end{tabular}
  \end{center}

\end{table}

\section{Limitations and future work}

In this work we have introduced the Satisfiability Transformer for SAT/UNSAT
problem classification and validated it on a subset of the SATComp
(SATComp) 2022 problem sets. While we have demonstrated a pure machine learning
solution that can handle SATComp-sized problems an order of magnitude larger
than previous work, evaluating all of SATComp would
require scaling our approach another order of magnitude and na\"ively doing so
would require a significant hardware commitment.

This work has also only demonstrated SAT/UNSAT classification, whereas
generation of satisfying variable assignments in the case of SAT, and the
UNSAT core in the case of UNSAT is not something that we consider
here. Predicting the satisfying variable assignment
and UNSAT core could be tackled with only minor modifications to our current
architecture and training scheme, as our data generation methods already yield
satisfying assignments and UNSAT cores.  Indeed, for SAT it is well known that
given a polynomial `oracle' for satisfiable formula, one can obtain an assignment quadratic
time (linear time in the time of evaluating the formula).  Suppose $\varphi$ is
satisfiable. Then either $\varphi[1/x_k]$ is satisfiable or not.  If so then the $k^\text{th}$
there is an assignment $\alpha$ with $\varphi(\alpha)=1$ and $\alpha_k=1$ otherwise
there is an $\alpha$ with $\varphi(\alpha)=1$ and $\alpha_k=0$.

Further, using a transformer
architecture to produce a sequential output, like a satisfying assignment or
UNSAT core, is a natural application for a transformer. Techniques from NLP for
sequence to sequence prediction, like masked language
modeling \cite{Devlin2019BERTPO}, could easily be brought to bear. We have
reported results on training with either the rows or columns of our variable
space encoding as tokens, we leave exploration on training with the combined set
of features to future work. We also leave a careful examination of the
interpretability of SaT to future work; in particular we suspect that one could
extract rule-based heuristics from the attention maps. Finally, a deeper theoretical understanding of how problem hardness
encountered during training impacts performance of learned predictors
remains a topic for future work.

\section{Conclusion}

We believe that by tackling hard problems such as SAT many insights
into both machine learning practice and the nature of these problems
can be gained. SAT and UNSAT are problems about a synthetic language with precise syntax and semantics
where we can actually provide ground truth about the statements involved that
are not open to interpretation. Many current problems involve some amount of
reasoning; isolating the learnability of this reasoning may lead to new insights
on bounds of what can be learned about reasoning in natural domains. The ideas
we have presented include immediate benefits: new approaches to fast, hardware
accelerated data generation for synthetic languages; new approaches to using
computer vision models in domains related to optimization; and an end-to-end
framework for learning hard problems in memory-efficient ways.

\section*{Acknowledgments}
This work greatly benefited from discussions with Aleksey Nogin on an
earlier version of this work.  We are also grateful to Cynthia Wang
for useful discussions of later versions of this and related work.

\bibliographystyle{abbrvnat}
\bibliography{atssh}

\begin{thebibliography}{39}
\providecommand{\natexlab}[1]{#1}
\providecommand{\url}[1]{\texttt{#1}}
\expandafter\ifx\csname urlstyle\endcsname\relax
  \providecommand{\doi}[1]{doi: #1}\else
  \providecommand{\doi}{doi: \begingroup \urlstyle{rm}\Url}\fi

\bibitem[Balyo et~al.(2022)Balyo, Heule, Iser, J\"{a}rvisalo, and
  (eds)]{satcompetition2022}
T.~Balyo, M.~J.~H. Heule, M.~Iser, M.~J\"{a}rvisalo, and M.~S. (eds).
\newblock Proceedings of {SAT} {C}ompetition 2022 : Solver and benchmark
  descriptions, 2022.
\newblock URL \url{https://helda.helsinki.fi/handle/10138/347211}.

\bibitem[Cook(1971)]{Cook1971TheCO}
S.~A. Cook.
\newblock The complexity of theorem-proving procedures.
\newblock \emph{Proceedings of the third annual ACM symposium on Theory of
  computing}, 1971.

\bibitem[Dai et~al.(2020)Dai, Lai, Yang, and Le]{Dai2020FunnelTransformerFO}
Z.~Dai, G.~Lai, Y.~Yang, and Q.~V. Le.
\newblock Funnel-transformer: Filtering out sequential redundancy for efficient
  language processing.
\newblock \emph{ArXiv}, abs/2006.03236, 2020.

\bibitem[Davis and Putnam(1960)]{article:dp-resolution}
M.~Davis and H.~Putnam.
\newblock A computing procedure for quantification theory.
\newblock \emph{Journal of the ACM}, 7\penalty0 (3):\penalty0 201--215, 1960.

\bibitem[Devlin and O'Sullivan(2008)]{Devlin08satisfiabilityas}
D.~Devlin and B.~O'Sullivan.
\newblock Satisfiability as a classification problem.
\newblock In \emph{PROC. OF THE 19TH IRISH CONF. ON ARTIFICIAL INTELLIGENCE AND
  COGNITIVE SCIENCE}, 2008.

\bibitem[Devlin et~al.(2019)Devlin, Chang, Lee, and
  Toutanova]{Devlin2019BERTPO}
J.~Devlin, M.-W. Chang, K.~Lee, and K.~Toutanova.
\newblock Bert: Pre-training of deep bidirectional transformers for language
  understanding.
\newblock \emph{ArXiv}, abs/1810.04805, 2019.

\bibitem[Dosovitskiy et~al.(2021)Dosovitskiy, Beyer, Kolesnikov, Weissenborn,
  Zhai, Unterthiner, Dehghani, Minderer, Heigold, Gelly, Uszkoreit, and
  Houlsby]{Dosovitskiy2021AnII}
A.~Dosovitskiy, L.~Beyer, A.~Kolesnikov, D.~Weissenborn, X.~Zhai,
  T.~Unterthiner, M.~Dehghani, M.~Minderer, G.~Heigold, S.~Gelly, J.~Uszkoreit,
  and N.~Houlsby.
\newblock An image is worth 16x16 words: Transformers for image recognition at
  scale.
\newblock \emph{ArXiv}, abs/2010.11929, 2021.

\bibitem[Feige and Jozeph(2012)]{conference:universalsatgraph}
U.~Feige and S.~Jozeph.
\newblock Universal factor graphs.
\newblock In A.~Czumaj, K.~Mehlhorn, A.~Pitts, and R.~Wattenhofer, editors,
  \emph{Automata, Languages, and Programming}, 2012.

\bibitem[Friedrich et~al.(2017)Friedrich, Krohmer, Rothenberger, Sauerwald, and
  Sutton]{friedrich_et_al:LIPIcs:2017:7835}
T.~Friedrich, A.~Krohmer, R.~Rothenberger, T.~Sauerwald, and A.~M. Sutton.
\newblock {Bounds on the Satisfiability Threshold for Power Law Distributed
  Random SAT}.
\newblock In K.~Pruhs and C.~Sohler, editors, \emph{25th Annual European
  Symposium on Algorithms (ESA 2017)}, volume~87 of \emph{Leibniz International
  Proceedings in Informatics (LIPIcs)}, pages 37:1--37:15, Dagstuhl, Germany,
  September 2017. Schloss Dagstuhl--Leibniz-Zentrum fuer Informatik.
\newblock ISBN 978-3-95977-049-1.
\newblock \doi{10.4230/LIPIcs.ESA.2017.37}.
\newblock URL \url{http://drops.dagstuhl.de/opus/volltexte/2017/7835}.

\bibitem[Gent and Walsh(1994)]{tech-report:gent-walsh-phase}
I.~Gent and T.~Walsh.
\newblock The sat phase transition.
\newblock Technical Report 679, University of Edinburgh, 1994.

\bibitem[Gholamalinezhad and Khosravi(2020)]{Gholamalinezhad2020PoolingMI}
H.~Gholamalinezhad and H.~Khosravi.
\newblock Pooling methods in deep neural networks, a review.
\newblock \emph{ArXiv}, abs/2009.07485, 2020.

\bibitem[Heo et~al.(2021)Heo, Yun, Han, Chun, Choe, and
  Oh]{Heo2021RethinkingSD}
B.~Heo, S.~Yun, D.~Han, S.~Chun, J.~Choe, and S.~J. Oh.
\newblock Rethinking spatial dimensions of vision transformers.
\newblock \emph{2021 IEEE/CVF International Conference on Computer Vision
  (ICCV)}, pages 11916--11925, 2021.

\bibitem[Hrube\u{s}(2007)]{article:intuitionistic-proof-complexity}
P.~Hrube\u{s}.
\newblock A lower bound for intuitionistic logic.
\newblock \emph{Annals of Pure and Applied Logic}, 146\penalty0 (1):\penalty0
  72--90, 2007.

\bibitem[Hrube\u{s}(2009)]{article:complexity-nonclassical}
P.~Hrube\u{s}.
\newblock On the lengths of proofs in non-classical logics.
\newblock \emph{Annals of Pure and Applied Logic}, 157:\penalty0 194--205,
  2009.

\bibitem[Huang et~al.(2022)Huang, Khetan, Bidart, and
  Karnin]{Huang2022PyramidBERTRC}
X.~Huang, A.~Khetan, R.~Bidart, and Z.~S. Karnin.
\newblock Pyramid-bert: Reducing complexity via successive core-set based token
  selection.
\newblock In \emph{Annual Meeting of the Association for Computational
  Linguistics}, 2022.

\bibitem[Impagliazzo(1995)]{article:impagliazzo-five-worlds}
R.~Impagliazzo.
\newblock A personal view of average-case complexity.
\newblock In \emph{Proceedinds of Structure in Complexity Theory}, volume~10,
  pages 134--147, 1995.

\bibitem[Kim and Cho(2020)]{Kim2020LengthAdaptiveTT}
G.~Kim and K.~Cho.
\newblock Length-adaptive transformer: Train once with length drop, use anytime
  with search.
\newblock In \emph{Annual Meeting of the Association for Computational
  Linguistics}, 2020.

\bibitem[Levin(2003)]{article:levin-oneway}
L.~A. Levin.
\newblock The tale of one-way functions.
\newblock \emph{Problems of Information Transmission}, 39:\penalty0 92--103,
  2003.

\bibitem[Lin et~al.(2016)Lin, Doll{\'a}r, Girshick, He, Hariharan, and
  Belongie]{Lin2016FeaturePN}
T.-Y. Lin, P.~Doll{\'a}r, R.~B. Girshick, K.~He, B.~Hariharan, and S.~J.
  Belongie.
\newblock Feature pyramid networks for object detection.
\newblock \emph{2017 IEEE Conference on Computer Vision and Pattern Recognition
  (CVPR)}, pages 936--944, 2016.

\bibitem[Lindenmayer(1968)]{LINDENMAYER1968300}
A.~Lindenmayer.
\newblock Mathematical models for cellular interactions in development ii.
  simple and branching filaments with two-sided inputs.
\newblock \emph{Journal of Theoretical Biology}, 18\penalty0 (3):\penalty0
  300--315, 1968.
\newblock ISSN 0022-5193.
\newblock \doi{https://doi.org/10.1016/0022-5193(68)90080-5}.
\newblock URL
  \url{https://www.sciencedirect.com/science/article/pii/0022519368900805}.

\bibitem[Megill and Wheeler(2019)]{metamath}
N.~D. Megill and D.~A. Wheeler.
\newblock \emph{Metamath: A Computer Language for Mathematical Proofs}.
\newblock Lulu Press, Morrisville, North Carolina, 2019.
\newblock {\tt http://us.metamath.org/downloads/metamath.pdf}.

\bibitem[Newsham et~al.(2015)Newsham, Lindsay, Ganesh, Liang, Fischmeister, and
  Czarnecki]{conference:satgraf}
Z.~Newsham, W.~Lindsay, V.~Ganesh, J.~H. Liang, S.~Fischmeister, and
  K.~Czarnecki.
\newblock Satgraf: Visualizing the evolution of sat formula structure in
  solvers.
\newblock In M.~Heule and S.~Weaver, editors, \emph{Theory and Applications of
  Satisfiability Testing}, 2015.

\bibitem[Pan et~al.(2021)Pan, Zhuang, Liu, He, and Cai]{Pan2021ScalableVT}
Z.~Pan, B.~Zhuang, J.~Liu, H.~He, and J.~Cai.
\newblock Scalable vision transformers with hierarchical pooling.
\newblock \emph{2021 IEEE/CVF International Conference on Computer Vision
  (ICCV)}, pages 367--376, 2021.

\bibitem[Pietruszka et~al.(2020)Pietruszka, Borchmann, and
  Garncarek]{Pietruszka2020SparsifyingTM}
M.~Pietruszka, {\L}.~Borchmann, and L.~Garncarek.
\newblock Sparsifying transformer models with trainable representation pooling.
\newblock In \emph{Annual Meeting of the Association for Computational
  Linguistics}, 2020.

\bibitem[Polu and Sutskever(2020)]{Polu2020GenerativeLM}
S.~Polu and I.~Sutskever.
\newblock Generative language modeling for automated theorem proving.
\newblock \emph{ArXiv}, abs/2009.03393, 2020.

\bibitem[Radford et~al.(2019)Radford, Wu, Child, Luan, Amodei, and
  Sutskever]{Radford2019LanguageMA}
A.~Radford, J.~Wu, R.~Child, D.~Luan, D.~Amodei, and I.~Sutskever.
\newblock Language models are unsupervised multitask learners.
\newblock In \emph{OpenAI}, 2019.

\bibitem[Samal et~al.(1994)Samal, Peterson, and Holliday]{413300}
A.~Samal, B.~Peterson, and D.~Holliday.
\newblock Recognizing plants using stochastic l-systems.
\newblock In \emph{Proceedings of 1st International Conference on Image
  Processing}, volume~1, pages 183--187 vol.1, 1994.
\newblock \doi{10.1109/ICIP.1994.413300}.

\bibitem[Selsam and
  Bj{\o}rner(2019{\natexlab{a}})]{10.1007/978-3-030-24258-9_24}
D.~Selsam and N.~Bj{\o}rner.
\newblock Guiding high-performance sat solvers with unsat-core predictions.
\newblock In M.~Janota and I.~Lynce, editors, \emph{Theory and Applications of
  Satisfiability Testing -- SAT 2019}, pages 336--353, Cham,
  2019{\natexlab{a}}. Springer International Publishing.
\newblock ISBN 978-3-030-24258-9.

\bibitem[Selsam and Bj{\o}rner(2019{\natexlab{b}})]{Selsam2019NeuroCoreGH}
D.~Selsam and N.~S. Bj{\o}rner.
\newblock Neurocore: Guiding high-performance sat solvers with unsat-core
  predictions.
\newblock \emph{ArXiv}, abs/1903.04671, 2019{\natexlab{b}}.

\bibitem[Selsam et~al.(2019)Selsam, Lamm, B{\"u}nz, Liang, de~Moura, and
  Dill]{Selsam2019LearningAS}
D.~Selsam, M.~Lamm, B.~B{\"u}nz, P.~Liang, L.~M. de~Moura, and D.~L. Dill.
\newblock Learning a sat solver from single-bit supervision.
\newblock \emph{ArXiv}, abs/1802.03685, 2019.

\bibitem[Shi et~al.(2022)Shi, Li, Khan, Zhen, jie Yuan, and
  Xu]{Shi2022SATformerTF}
Z.~Shi, M.~Li, S.~Khan, H.-L. Zhen, M.~jie Yuan, and Q.~Xu.
\newblock Satformer: Transformers for sat solving.
\newblock \emph{ArXiv}, abs/2209.00953, 2022.

\bibitem[Tseitin(1968)]{article:tseitin-transform}
G.~S. Tseitin.
\newblock On the complexity of derivations in the propositional calculus
  (translated from russian).
\newblock In A.~O. Slisenko, editor, \emph{Studies in constructive mathematics
  and mathematical logic, part II}, pages 115--125. Springer, 1968.

\bibitem[Wang et~al.(2021{\natexlab{a}})Wang, Hu, Tiwari, Khurshid, McMillan,
  and Miikkulainen]{Wang2021NeuroCombIS}
W.~Wang, Y.~Hu, M.~Tiwari, S.~Khurshid, K.~L. McMillan, and R.~Miikkulainen.
\newblock Neurocomb: Improving sat solving with graph neural networks.
\newblock \emph{ArXiv}, abs/2110.14053, 2021{\natexlab{a}}.

\bibitem[Wang et~al.(2021{\natexlab{b}})Wang, Xie, Li, Fan, Song, Liang, Lu,
  Luo, and Shao]{Wang2021PyramidVT}
W.~Wang, E.~Xie, X.~Li, D.-P. Fan, K.~Song, D.~Liang, T.~Lu, P.~Luo, and
  L.~Shao.
\newblock Pyramid vision transformer: A versatile backbone for dense prediction
  without convolutions.
\newblock \emph{2021 IEEE/CVF International Conference on Computer Vision
  (ICCV)}, pages 548--558, 2021{\natexlab{b}}.

\bibitem[Wen and Yu(2023)]{DBLP:journals/corr/abs-2302-00272}
W.~Wen and T.~Yu.
\newblock {W2SAT:} learning to generate {SAT} instances from weighted literal
  incidence graphs.
\newblock \emph{CoRR}, abs/2302.00272, 2023.
\newblock \doi{10.48550/arXiv.2302.00272}.
\newblock URL \url{https://doi.org/10.48550/arXiv.2302.00272}.

\bibitem[Wu et~al.(2021)Wu, Liu, and Liu]{Wu2021CentroidTL}
L.~Wu, X.~Liu, and Q.~Liu.
\newblock Centroid transformers: Learning to abstract with attention.
\newblock \emph{ArXiv}, abs/2102.08606, 2021.

\bibitem[Xu et~al.(2007)Xu, Hutter, Hoos, and
  Leyton-Brown]{10.1007/978-3-540-74970-7_50}
L.~Xu, F.~Hutter, H.~H. Hoos, and K.~Leyton-Brown.
\newblock Satzilla-07: The design and analysis of an algorithm portfolio for
  sat.
\newblock In C.~Bessi{\`e}re, editor, \emph{Principles and Practice of
  Constraint Programming -- CP 2007}, pages 712--727, Berlin, Heidelberg, 2007.
  Springer Berlin Heidelberg.
\newblock ISBN 978-3-540-74970-7.

\bibitem[Ye et~al.(2021)Ye, Lin, Huang, and Sun]{Ye2021TRBERTDT}
D.~Ye, Y.~Lin, Y.~Huang, and M.~Sun.
\newblock Tr-bert: Dynamic token reduction for accelerating bert inference.
\newblock In \emph{North American Chapter of the Association for Computational
  Linguistics}, 2021.

\bibitem[Zhao et~al.(2016)Zhao, Shi, Qi, Wang, and Jia]{Zhao2016PyramidSP}
H.~Zhao, J.~Shi, X.~Qi, X.~Wang, and J.~Jia.
\newblock Pyramid scene parsing network.
\newblock \emph{2017 IEEE Conference on Computer Vision and Pattern Recognition
  (CVPR)}, pages 6230--6239, 2016.

\end{thebibliography}

\appendix
\section{SAT generator proofs} \label{appendix:proof:sat-soundness}
In this section, we provide a proof of Proposition \ref{proposition:sat-soundness} and we also
provide some exhibition of additional properties enjoyed by the SAT generator.

We stay that a literal $l$ is \emph{underlied} by a variable $x$ if $l=x$ or $l=\neg x$.

\begin{proposition*}[Proposition \ref{proposition:sat-soundness}]
  \Paste{proposition:sat-soundness}
\end{proposition*}
Before beginning the proof, we say that a literal $l$ \emph{agrees} with an assignment $\alpha=(\alpha_0,\ldots,\alpha_n)$ if
$l(\alpha)=\mathsf{True}$.  Let the variable underlying $l$ be $x_j$,  the agreement
of $l$ with $\alpha$ can be made on cases; if $l=x_j$ is a positive variable then $\alpha_j=1$, and if $l=\neg x_j$
is a negative variable then $\alpha_j=0$.

\begin{proof}
  Suppose Algorithm \ref{algorithm:satgen} is run with inputs $n,m,\mathsf{var}$, $\mathsf{lits\mbox{-}clause}$, $\mathsf{polarities}$,
  and we obtain $\varphi$ as output.  To show that $\varphi$ is satisfiable, we must exhibit
  an assignment $\alpha$ with $\varphi(\alpha) = \mathsf{True}$.  Since $\varphi$ is
  a CNF, it then suffices to show that for each clause $c_i$, $c_i(\alpha) = \mathsf{True}$.
  Then it suffices to find some literal $l_{i0}$ of $c_i$ which agrees with $\alpha$ for each $i$.

  Note that $\mathsf{asst}$ generated in the first step of the algorithm gives us such an assignment, by changing all the occurrences of $-1$ to $0$.
  To see this, the inner for loop sets the $j^\text{th}$ variable of $c_i$ to agree with $\alpha_j$ if the
  $j^\text{th}$ element of seq is a $1$.  Then we are guaranteed that one literal agrees with $\alpha$ as long as
  $\mathsf{seq}$ contains at least one $1$; however, $\mathsf{seq}$ is chosen from a distribution of
  non-empty strings that contain at least one $1$.
\end{proof}

Note that the satisfiable instance generator, Algorithm \ref{algorithm:satgen}, is parameterized over four distributions;
the vars distribution determines how likely each variable is to appear in the formula, the lits-clause distribution
determines the likelihood of literals-per-clause in the formula, the polarities distribution determines a bias
over $0,1$ on the satisfying assignment that the formula will satisfy, and the polarity-bias determines a bias
over how many literals in the clause will agree with the satisfying assignment (at least one must).  These then allow
generating formulae with a variety of distributional phenomena, for example completeness.

Then,
\begin{lemma}\label{lemma:absolute-sat}
  Suppose $\mathsf{vars}$ is the uniform distribution on $m$ characters, $\mathsf{lits\mbox{-}clause}$ is
  the constantly $k$ distribution, $\mathsf{polarities}$ is the Bernoulli distribution with parameter $p=0.5$ and
  $\mathsf{polarity\mbox{-}bias}$ is the uniform distribution on the $2^{k}-1$ sequences of $\{0,1\}^k$ that contain at
  least one $1$.  Then $\mathsf{GenerateSAT}(n,m,\mathsf{vars},\mathsf{lits\mbox{-}clause},\mathsf{polarities},\mathsf{polarity\mbox{-}bias})$
  will eventually generate every $n$-clause, $m$ variable, satisfiable $k$-CNF.
\end{lemma}
\begin{proof}
  Fixing the distributions in the above, denote the distribution on $n$-clause, $m$-variable CNF produced by $\mathsf{GenerateSAT}$ by
  \[GS(n,m):= \mathsf{GenerateSAT}(n,m,\mathsf{vars},\mathsf{lits\mbox{-}clause},\mathsf{polarities},\mathsf{polarity\mbox{-}bias})\]
  entirely.  Note that we can then refactor $GS(n,m)$ into step 1 where sample an assignment $\alpha$ and the rest of the
  steps $GS_0(\alpha,n,m)$ which creates a formula that satisfies $\alpha$.  Let $\varphi \leftarrow GS_0(\alpha,n,m)$.

  One may note then, that each clause in $\varphi$ is sampled uniformly over all $k$-clauses in $m$-variables that satisfy
  $\alpha$.  To see this, just note that $\mathsf{polarity\mbox{-}bias}$ selects uniformly from all possible $2^k-1$
  ways $k$ variables to satisfy the $\alpha$, and $\mathsf{vars}$ samples uniformly from the $m$ variables that will
  appear in the clause.

  It then follows that, $GS_0(\alpha,n,m)$ is the uniform distribution over all satisfiable $n$-clause, $m$-variable, $k$-CNF,
  conditional on satisfying $\alpha$.

  Since we then sampled $\alpha$ uniformly and with no polarity bias, $\alpha$ is uniform over length $m$ assignments.

  It then follows from Bayes' theorem that $GS$ is distributed uniformly over pairs $(\varphi,\alpha)$  such that $\varphi(\alpha)=1$.

  Now note that any satisfiable formula must have some $\alpha$ with $\varphi(\alpha)=1$, so that eventually generating
  $\varphi$ is implied by eventually generating every $(\varphi,\alpha)$ with $\varphi(\alpha)=1.$  But $GS$ is uniform,
  hence will eventually generate every such pair, hence completing the proof.
\end{proof}

We leave for future work some of the more interesting and nuanced distributional phenomena.  For example,
in the above, we conditioned on a single $\alpha$, but realistically, one might want to condition
on arbitrary sets of assignments, and pinning down the exact relationship between the distribution conditioned on
arbitrary sets of assignments and the one used above requires additional work.  This kind of more nuanced
view also allows other analyses that compare directly to the distribution of all satisfiable formula.
We can also consider non-uniform distributions, such as the power-law distribution on $\mathsf{vars}$, and
how this affects generated formulae.  One interesting feature of such analyses captures the difference
between the biased distribution of CNF and the biased distribution of satisfiable CNF.  For example, even
in the uniform case, some of the classical phase transition \cite{tech-report:gent-walsh-phase} work
reveals different phase transitions in hardness for satisfiable formula given satisfiability versus all formula.

\section{UNSAT generator proofs}\label{appendix:proof:unsat-soundness}
We call a clause \emph{full} when each of the $n$ variables appears in it.

\begin{lemma*}\label{lemma:ressearch-pres-unsat}
  Suppose $\mathsf{\sc ResSearch}$ is called on $\mathsf{prob}_1,\mathsf{vars},\mathsf{bloom}$, with $\mathsf{prob}_1 \wedge \mathsf{prob}_2$
  unsatisfiable and where $\mathsf{prob}_1$ has no full clauses.  Denote by $\mathsf{prob}_1'$ the returned problem.  Then
  $\mathsf{prob}_1' \wedge \mathsf{prob}_2$ is unsatisfiable.
\end{lemma*}
Here we use that a formula $\varphi$ is unsatisfiable iff $\varphi \Rightarrow \bot$.
\begin{proof}
  Suppose we call $\mathsf{\sc ResSearch}$ on $\varphi,\mathsf{vars},\mathsf{bloom}$ with $\varphi$
  containing no full clauses, and obtain the result $\varphi'$.  Since $\varphi$ has no full-clauses,
  every clause will split under $\mathsf{ResSearch}$.

  $\varphi'$ is a formula in $2m$ clauses when $\varphi$ is a formula in $m$ clauses.
  Denote the $t^\text{th}$ clause of $\varphi,\varphi'$ by $c_t,c_t'$ respectively.
  Then we will show that for $i \in \{1,\ldots,m\}$, $c'_{2i} \wedge c'_{2i-1} \Rightarrow c_i$;
  hence $\varphi' \Rightarrow \varphi$.

  Suppose $c_i$ has $k_i$ literals and write $c_i=\vee_{j=1}^{k_i} l_{ij}$.
  Then the algorithm samples $k_i$ elements from $\mathsf{bloom}$, $b_1,\ldots,b_{k_i}$ which
  determine whether $l_{ij}$ appears in $c'_{2i},c'_{2i-1}$ or both.  Let
  $\overline{l_{ij}}^{1}$ be $l_{ij}$ if $b_j=0,2$ else $0$ and $\overline{l_{ij}}^{2}$ be $l_{ij}$ if
  $b_j=1,2$ else $0$.  Since $b_j$ is sampled from a distribution over $\{0,1,2\}$ at least one
  of $\overline{l_{ij}}^{1},\overline{l_{ij}}^{2}$ is $l_{ij}$ and hence moreover $\overline{l_{ij}}^{1} \vee \overline{l_{ij}}^{2} = l_{ij}$.

  The algorithm constructs $c'_{2i},c'_{2i-1}$ by choosing an unused variable, $x_0$, and then setting:
  \[
    c'_{2i}=\left(\vee_{j=1}^{k_i}\overline{l_{ij}}^{1} \right)\vee x_0
    \qquad
    c'_{2i-1}=\left( \vee_{j=1}^{k_i}\overline{l_{ij}}^{2} \right) \vee \neg x_0
  \]

  Note that an unused variable $x_0$ is guaranteed to exist since $\varphi$ contains no full clauses.
  Also, no $l_{ij}$ is underlied by $x_0$, hence the $x_0$
  appearing in the above is neither absorbed by idempotency nor canceled by the law of the excluded middle.
  Thus, we may apply a contraction of the resolution rule:
  \[
    c'_{2i}\wedge c'_{2i-1}
    \Rightarrow \left(\vee_{j=1}^{k_i}\overline{l_{ij}}^{1} \right)\vee\left( \vee_{j=1}^{k_i}\overline{l_{ij}}^{2} \right)
    \equiv \vee_{j=1}^{k_i} \left(\overline{l_{ij}}^1 \vee \overline{l_{ij}}^2\right)
    \equiv \vee_{j=1}^{k_i} \left(l_{ij}\right)
    \equiv c_i
  \]
  as desired.  Thus we've shown that $\varphi' \Rightarrow \varphi$.

  Suppose we call $\mathsf{\sc ResSearch}$ on $\varphi,\mathsf{vars},\mathsf{bloom}$ with $\varphi$
  unsatisfiable and containing no full clauses, and obtain the result $\varphi'$.  Since $\varphi$ is unsatisfiable, we have that $\varphi \Rightarrow \bot$
  and it suffices to show that $\varphi' \Rightarrow \varphi$ since then by transitivity we would have that $\varphi'\Rightarrow \bot$;hence,
  unsatisfiable.  In fact we will show that $\varphi' \Rightarrow \varphi$ by a parallel
  resolution step.

  Now suppose that $\varphi_1\wedge \varphi_2$ is unsatisfiable; i.e., $\varphi_1 \wedge \varphi_2 \Rightarrow \bot$, and that $\varphi_1$ has no full clauses.
  Suppose we obtain $\varphi_1'$ by running $\mathsf{ResSearch}$ on $\varphi_1$.  Then since conjunction preserves implication and by transitivity of implication we have
  $\varphi_1' \wedge \varphi_2 \Rightarrow \varphi_1 \wedge \varphi_2 \Rightarrow \bot$, whence, $\varphi_1' \wedge \varphi_2$ is unsatisfiable as required.
\end{proof}

\begin{proposition*}[Proposition \ref{proposition:unsat-soundness}]
  \Paste{proposition:unsat-soundness}
\end{proposition*}
Again we use that a formula $\varphi$ is unsatisfiable iff $\varphi \Rightarrow \bot$.
\begin{proof}
  Suppose we run algorithm \ref{algorithm:unsatgen} with inputs
  $n$, $m$, $\mathsf{init\mbox{-}size}$, $\mathsf{depth}$,
  $\mathsf{vars}$, $\mathsf{lits\mbox{-}clause}$,
  $\mathsf{polarities}$, $\mathsf{bloom}$,
  with $\mathsf{init\mbox{-}size} > 0$.

  Note that the algorithm returns a formula of the form $\varphi \wedge \psi$ where
  $\psi$ is randomly generated.  It then suffices to show that $\varphi$ is unsatisfiable since
  $\varphi \wedge \psi \Rightarrow \varphi$.

  The formula $\varphi$ is constructed in two phases.  In the first phase we initialize a formula
  as pairs of formulas of the form $x\wedge \neg x$; i.e., pairs of contradictions hence unsatisfiable.
  We then iteratively update the problem by applying $\mathsf{\sc ResSearch}$ to create a new problem from the current problem,
  and $\varphi$ is the problem that results from this iterative procedure.

  Note that if the formula created in the initialization phase has a full clause (i.e. $n=1$), then the
  loop immediately exits, and $\varphi$ is the formula of pairs of contradictions; hence unsatisfiable.
  Otherwise we apply the loop at least once, and it suffices to show that $\mathsf{\sc ResSearch}$
  preserves unsatisfiability for formulae with no full clauses. Lemma \ref{lemma:ressearch-pres-unsat}
  can be applied to conclude the proof.
\end{proof}

Stating the sense in which the the unsatisfiable generator is complete is more subtle than for the satisfiable
generator.  The awkward bit is that we allow, for efficiency reasons, applying blooms in parallel.  However, since
we only bloom on a subset of the problem at each iteration, this blooming process can be simulated by one which
blossoms one step at a time, allowing us to relate the resulting formula to the usual notion of length; namely, the number of
resolution steps required to reach absurdity in some unsatisfiable core.  We can make this precise by
restricting our generator in a few ways.

\begin{lemma}
  Suppose we restrict $\mathsf{down\mbox{-}clause}$ so that it samples exactly one clause per round, with $\mathsf{init\mbox{-}size}$ the distribution concentrated in $2$,
  and so that $\mathsf{bloom}(u)$ is uniform over $\{0,1,2\}^{u}$, and where $\mathsf{lits\mbox{-}clause},\mathsf{vars},\mathsf{polarities}$ are all uniform.
  Then $\mathsf{GenerateUNSAT}$ of Algorithm \ref{algorithm:unsatgen} will eventually generate every
  unsatisfiable formula with a resolution core that admits some resolution proof whose length is linear in the size of the formula.
\end{lemma}
\begin{proof}
  With the above restrictions, the lemma follows immediately.  Note that every unsatisfiable CNF $\psi$ is of the form
  $\psi \equiv \varphi_c \wedge \varphi$ where $\varphi_c$ is an unsatisfiable core, that is some subformula that is unsatisfiable formula and $\varphi$ is
  arbitrary.  Moreover, every such $\psi$ can be put in the form $\psi \equiv \varphi_c \wedge \varphi$ where there is a chain of implications
  $\varphi_c \Rightarrow \varphi_{c,1} \Rightarrow \cdots \Rightarrow \varphi_{c,J} \Rightarrow \bot$ and each of the proof steps requires all the clauses in the
  preceding step (note that here we are talking about syntactic choice of proof step (since we need to have a length in some proof system), and not any proof that $\varphi_{c,i} \Rightarrow \varphi_{c,i+1}$).

  $\mathsf{GenerateUNSAT}$ generates precisely formulae of the form $\varphi_c \wedge \varphi$ where $\varphi_c$ is an unsatisfiable core and
  $\varphi$ is uniformly generated over all CNF, and hence we eventually generate all such $\varphi$.  Thus, by the above we are done
  if we can prove that we generate all $\varphi_c$ that are unsatisfiable core whose resolution length is linear in the size size of $\varphi_c$.

  However, this is by construction with the restriction on $\mathsf{down\mbox{-}clause}$ and $\mathsf{bloom}$.  We start with $\bot$, and
  we construct $\varphi_c$ one resolution step at a time, by choosing uniformly over all possible one-step backwards resolution steps.
  Thus we eventually generate every possible sequence of such moves.
\end{proof}

We will leave refinements and further analysis of the unsatisfiable instance generator for future work.  First, one may wish to choose
more interesting distributions over bloom that lead to unbalanced, more `real-world' looking trees; for example following alternate algorithms
and distributions for expanding $L$-systems.

Secondly, one may wish to use more interesting proof systems than resolution.  We chose
resolution here as a starting point for generating unsatisfiable problems.  For example,
one might wish to explore a proof system that would generate all unsatisfiability cores.
However, open problems present a subtlety in doing so efficiently.  For instance,
it is well known that a resolution proof of the pigeonhole principle requires exponentially many steps in the size of
formula \cite{Cook1971TheCO}.  Thus, our generator in its current form is not complete for all unsatisfiable formulae (since we generate
linear-length proofs).  Also, given the exponential length proof size of resolution proofs in general, we wouldn't want to base a
generator off resolution anyways.  The nice aspect about resolution is that it is refutationally complete; that is, given any proposition of the form
$\neg \varphi$, a resolution proof exists.  A slightly stronger notion is intuitionistically complete; these are proof systems
that prove any theorem in intuitionistic logic; note that as refutations, $\neg \varphi$ satisfy double negation, intuitionistically, i.e. $\neg\neg\neg \varphi \Rightarrow \neg \varphi$,
any intuitionistically complete proof system is refutation complete.  Alas, there are known formulae families of intuitionistic tautologies
that require $2^{n^{\Omega(1)}}$ intuitionistic Frege proof steps in any Frege system \cite{article:intuitionistic-proof-complexity,article:complexity-nonclassical}.
However, for classical Frege systems, there are no known superlinear bounds on proof size;
this problem is believed to be hard -- indeed a proof that every theorem has a polynomial
sized proof in a classical Frege system would suffice to prove NP=coNP.  It would indeed be
interesting to explore generating unsatisfiable formula using either intuitionistic or classical
short Frege proofs.  However, due to the open problems in complexity theory, either we know
such a generator will not be complete (in the intuitionistic case) or that completeness would require
a proof of a quite hard theorem (in the classical case).  Again, from the perspective of machine
learning, there is no point in generating exponential sized proofs, since this would make the
generator unbearably slow.  So while it seems that there might not be a provably correct way to
efficiently generate all unsatisfiable cores; it does not rule out that using more interesting
proof systems than resolution would yield interesting results.

\section{Hyperparameters}\label{appendix:hyperparameters}
\subsection{SaT training}
We experimented with various hyperparameters during training, though we held many
of the model hyperparameters constant across problem sizes. In particular, we
use a learning rate of 1e-4, an embedding dimension of 32, 8 attention heads,
prepend 31 \texttt{[concept]} tokens and 1 \texttt{[class]} token, and always
perform head slicing after the first layer and keep only the first 32 output
tokens.

Our hyperparameters for SaT are informed by our desire to run our data
generation and model training regime on a single NVIDIA Tesla V100S GPU, though
in practice we distribute training across 4 GPUs and average the gradients in
order to update the model. We realize our data generators and SaT in PyTorch and
utilize the Distributed Data Parallel module for multi-GPU training.

For the problem size encountered during training, we select the number of variables uniformly from $4$ up to
a maximum number of variables being trained on.  The minimum number of variables is chosen so that with the minimum
clause to variable ratio, we can always generate unsatisfiable problems.  The number of clauses selected depends on hyperparameters that
determine the clause to variable ratio.  These hyperparameters are shown in Table \ref{table:c-to-v-hypers}.
\begin{table}
  \caption{Clause to variable ratio hyperparameters}
  \smallskip
  \centering
  \begin{tabular}{llll}
    \toprule
    \cmidrule(r){1-3}
    Clause to & Generator  & Standard & Clip \\
    Variable Ratio & Option & Deviation & \\
    \midrule
    $c_\phi$ \cite{tech-report:gent-walsh-phase} & Uniform Mixed $k$-CNF & 1.0 & (2,11)\\
    3.71 & powerlaw $3$-CNF with exponent 2.6 \cite{friedrich_et_al:LIPIcs:2017:7835} & 1.0 & (2,11)\\
    4.27 & all other options & 1.0 & (2,11)\\
  \end{tabular}
  \label{table:c-to-v-hypers}
\end{table}

The algorithm for producing a satisfiable formula can be modified slightly to turn any formula $\varphi$
into a new formula $\varphi'$ with the same literal incidence graph
that is guaranteed to be satisfied by a chosen
assignment.  We realize this  modification by skipping directly to the assignment of polarities, taking
$\mathsf{var\mbox{-}indices}$ to be determined by the literal incidence graph of the starting problem.
There is also a more biased algorithm for turning a formula into a satisfiable one with the
same literal incidence graph.  In this algorithm, we select one literal
per clause and, if needed, change it so that it agrees with the
assignment. Yet another related method
proceeds in the same way, but also randomly flips the polarity of another literal each time the first chosen literal is flipped.
This maintains both the literal incidence graph and the count of positive and negative polarities.  We call the first option
the biased SAT cover and the second option the biased SAT cover with flip.

For example, applying the first of these procedures to a $3$-CNF generated with a uniform $\mathsf{polarity\mbox{-}bias}$ is equivalent to producing CNF
with a $\mathsf{polarity\mbox{-}bias}$ with the probabilities of agreement
are: exactly one literal: $0.25$; exactly two literals: $0.5$; exactly three literals: $0.25$.

These procedures provide alternate ways to generate satisfiable formulae that have a
polarity distribution that matches the unsatisfiable generator, which can be subtle due to the interaction of
the $\mathsf{bloom}$ distribution, the depth to which we bloom, and the conditions under which
we break out of the while-loop in Algorithm \ref{algorithm:unsatgen}.

We may also modify the unsatisfiable generator to conjunct not with a randomly generated formula but
instead a satisfiable formula.  This lets us generate unsatisfiable problems with feature distributions that
match the satisfiable generator.

Finally we also select which distribution we use for the variable, literals-per-clause, and bloom distributions randomly from a
weighted list of options.

We then select the generator option from a weighted list of options, and these weights are hyperparameters of training
which we include in Table \ref{table:generator-distribution}.  In this table use the phrase `+ distribution shift' to
denote that the option selects from the distributions of variables, literals-per-clause, and blooming.
\begin{table}
  \caption{Distribution of SAT and UNSAT generator choices utilized during training.}
  \smallskip
  \centering
  \begin{tabular}{lll}
    \toprule
    \cmidrule(r){1-3}
    Generator option & SAT/UNSAT & Weight \\
    \midrule
    Uniform $\mathsf{polarity\mbox{-}bias}$  & SAT & 41\% \\
    Biased cover with flip & SAT & 1\% \\
    Biased cover without flip & SAT & 1\% \\
    From random with uniform $\mathsf{polarity\mbox{-}bias}$ & SAT & 5\% \\
    \hspace{0.5cm} + distribution shift & & \\
    From UNSAT Uniform $\mathsf{polarity\mbox{-}bias}$ & SAT & 5\%
    \\
    \hspace{0.5cm} + distribution shift & & \\
    Uniform $\mathsf{polarity\mbox{-}bias}$&
    SAT & 20\% \\
    \hspace{0.5cm} + distribution shift & & \\
    From random via biased cover with &
    SAT & 5\% \\
    \hspace{0.5cm} flip + distribution shift &  & \\
    From random via biased cover without &
    SAT & 5\% \\
    \hspace{0.5cm}flip + distribution shift  & & \\
    From  UNSAT via biased cover  &
    SAT & 6\% \\
    \hspace{0.5cm} with flip + distribution shift & & \\
    From UNSAT via biased cover & SAT
    & 6\% \\
    \hspace{0.5cm}with flip + distribution shift  & & \\
    From SAT via biased cover  & SAT &
    5\% \\
    \hspace{0.5cm}with flip + distribution shift & & \\
    \midrule
    Shallow bloom  & UNSAT & 43\% \\
    Deep bloom& UNSAT & 10\% \\
    Shallow bloom +  distribution shift & UNSAT & 31\% \\
    Deep bloom  + distribution shift & UNSAT & 5\% \\
    From SAT via shallow bloom + distribution shift & UNSAT & 10\% \\
    From SAT via deep bloom + distribution shift & UNSAT & 1\% \\
    \bottomrule
  \end{tabular}
  \label{table:generator-distribution}
\end{table}

Then we select the distributions in the case of `+ distribution shift'  by a weighted list on the choices
of distributions, and these too, become hyperparameters of training.
The weights on the choices of variable distribution, literals-per-clause, polarity, polarity bias, and bloom are also hyperparameters of training
and can be found in Table \ref{table:distribution-distribution}.

\begin{table}
  \caption{Distribution of distributions utilized during training.}
  \smallskip
  \centering
  \begin{tabular}{llll}
    \toprule
    \cmidrule(r){1-3}
    Distribution & Use & Parameters & Weight \\
    \midrule
    Uniform & $\mathsf{vars}$ & (low=0,high=max variables) & 70\% \\
    Pareto & $\mathsf{vars}$ & (shape=1.16,scale=2) & 20\% \\
    Power-law & $\mathsf{vars}$ & $\beta=2.6$ & 0\% \\
    Log-normal &$\mathsf{vars}$ & (10,2) & 10\% \\
    \midrule
    Normal & $\mathsf{lits\mbox{-}per\mbox{-}clause}$ & (4.5,1.0) & 90\% \\
    Uniform & $\mathsf{lits\mbox{-}per\mbox{-}clause}$  & (3,7) & 10\%\\
    \midrule
    Bernoulli & $\mathsf{polarities}$ & 0.5 & 80\% \\
    Bernoulli & $\mathsf{polarities}$  & 0.3 & 10\%\\
    Bernoulli & $\mathsf{polarities}$ & 0.7& 10\%\\
    \midrule
    Uniform & $\mathsf{polarity\mbox{-}bias}$ & n/a& 100\% \\
    $k-1$-bias & $\mathsf{polarity\mbox{-}bias}$ & n/a & 0.0 \%\\
    \midrule
    Weights & $\mathsf{bloom}$ & (0.48,0.48.02) & 85\%\\
    Weights & $\mathsf{bloom}$ & (0.5,0.3,0.2) & 15\%\\
    \bottomrule
  \end{tabular}
  \label{table:distribution-distribution}
\end{table}

Except for the
cases where we are comparing to NeuroSAT, the selection
of which generator option is applied is found in table \ref{table:generator-distribution}.

For the comparison to NeuroSAT, we are isolating the effect of the polarity bias distribution,
so the different experiments use different distributions.

\end{document}